\documentclass[10pt,journal,compsoc]{IEEEtran}
\ifCLASSOPTIONcompsoc
  \usepackage[nocompress]{cite}
\else
  \usepackage{cite}
\fi
\ifCLASSINFOpdf
  \usepackage[pdftex]{graphicx}
\else
\fi
\usepackage{amsmath}
\newcommand{\etal}{\textit{et al.}}
\usepackage{color}
\newcommand{\SAHA}[1]{\textcolor{black}{#1}}
\usepackage{booktabs} 
\usepackage{hyperref}
\begin{document}
%

\title{Two-Stream AMTnet for Action Detection}

\author{Suman~Saha$^{*}$,~Gurkirt~Singh$^{*}$~and~Fabio~Cuzzolin
\IEEEcompsocitemizethanks{\IEEEcompsocthanksitem F. Cuzzolin
is with the Visual Artificial Intelligence Laboratory, School of Engineering, Computing and Mathematics, Oxford Brookes University, United Kingdom.
S. Saha and G.Singh are with the Computer Vision Laboratory, ETH Zurich, Switzerland.
(*) S. Saha and G. Singh have contributed to this paper equally.
E-mail: (suman.saha,gurkirt.singh)@vision.ee.ethz.ch, fabio.cuzzolin@brookes.ac.uk}
\thanks{Manuscript received XXX, 2020.}}

\IEEEtitleabstractindextext{%
\begin{abstract} 
In this paper, we propose a new deep neural network architecture for online action detection, termed \emph{Two-Stream AMTnet},
which leverages recent advances in video-based action representation~\cite{saha2017amtnet} and incremental action tube generation~\cite{singh2017online}.
Majority of the present action detectors follow a frame-based representation, a late-fusion followed by an offline action tube building steps.
These are sub-optimal as: frame-based features barely encode the temporal relations; late-fusion restricts the network to learn robust spatiotemporal features; and finally, an offline action tube generation is not suitable for many real world problems such as autonomous driving, human-robot interaction to name a few.
\SAHA{The key contributions of this work are: 
(1) combining AMTnet's 3D proposal architecture with an online action tube generation technique
which allows the model to learn stronger temporal features needed for accurate action detection and 
facilitates running inference online (i.e. incrementally build action tubes in time);
(2) an efficient fusion technique allowing the deep network to learn strong spatiotemporal action representations.}
This is achieved by augmenting the previous Action Micro-Tube (AMTnet) action detection framework in three distinct ways:
by adding a parallel motion stream to the original appearance one in AMTnet;
(2) in opposition to state-of-the-art action detectors which train appearance and motion streams separately, and use a test time late fusion scheme to fuse RGB and flow cues,
by jointly training both streams in an end-to-end fashion and merging RGB and optical flow features at training time, thus allowing the network to learn pixel-wise correspondences between appearance and motion features;
(3) by introducing an online action tube generation algorithm which works at video-level (as opposed to frame-level~\cite{singh2017online}),
and in real-time (when exploiting only appearance features).
Two-Stream AMTnet exhibits superior action detection performance over state-of-the-art approaches on the standard action detection benchmarks UCF-101-24 and J-HMDB-21.
  
\end{abstract}

}

\maketitle

\IEEEdisplaynontitleabstractindextext

%
\IEEEpeerreviewmaketitle


\IEEEraisesectionheading{\section{Introduction}\label{sec:intro}}

\IEEEPARstart{S}{patiotemporal} human action detection in videos is one of the core challenging problems in computer vision.
Despite the notable progress made in this area in recent years, however, current state of the art approaches~\cite{Georgia-2015a,Weinzaepfel-2015,peng2016eccv,Saha2016,kalogeiton2017action,hou2017tube} share a number of limitations.
\textbf{(a)} Firstly, they adopt a \emph{frame-based} feature representation (except~\cite{kalogeiton2017action,hou2017tube}) suitable for image-level detection tasks such as object detection in images~\cite{girshick2016region,girshick2015fast}. This limits these models to learn only spatial or appearance-based features, 
and prevents these architectures from taking full advantage of the temporal information~\cite{saha2017amtnet} carried by videos.
\textbf{(b)} Secondly, current methods train `appearance' (derived from RBG information) and `motion' (based on optical flow calculation) streams separately, and fuse the related cues at \emph{test time}. This hinders these models from learning the complementary contributions of appearance and motion to class discrimination at training time, as they fail  
to provide a model of how these cues evolve over time~\cite{Feichtenhofer_2016_CVPR}. 
As two individual networks for appearance and motion stream are trained independently,  optimisation is run \emph{disjointly}, preventing these frameworks to be trained in an {end-to-end} fashion.
\textbf{(c)} As a third limitation, several such action detectors~\cite{Georgia-2015a,Weinzaepfel-2015,peng2016eccv,Saha2016} are only suitable for \emph{offline} processing, i.e., they assume that the entire video (taken as a 3D block of pixels) is available ahead of time in order to detect action instances.
Although \cite{kalogeiton2017action,hou2017tube} use an online algorithm for generating `action tubes' (i.e., sequences of frame-level detections corresponding to an action instance), they are  limited by their stacking $k$ video frames at each time instant,
which is expensive and unsuitable for real-time applications.
For an action detector to be applicable to real-world scenarios such as video surveillance and human-robot interaction, instead, video frames need to be processed \emph{online} -- 
action tubes need to be generated incrementally over time as the video streams in.
Moreover, the test time {detection speeds} of these detectors are too low to be suitable for real-world deployment.
\textbf{(d)} Finally, action detection methods such as~\cite{kalogeiton2017action,hou2017tube} require a source of dense ground truth for network training. Annotation in the form of bounding boxes is required for $k$ consecutive video frames in any given training example --
Kalogeiton~\etal~\cite{kalogeiton2017action} use $k=6$ whereas for Hou~\etal~\cite{hou2017tube} $k=8$.
Generating such dense bounding box annotation for long video sequences is highly expensive and impractical~\cite{daly2016weinzaepfel,ava2017gu}.
The latest generation action detection benchmarks DALY~\cite{daly2016weinzaepfel} and AVA~\cite{ava2017gu}, in contrast, only consider more realistic, sparser annotations.
More specifically, DALY provides $1$ to $5$ frames annotated with bounding boxes per action instance, irrespective of its duration, whereas AVA comes with only one annotated frame per second.
\begin{figure*}[!t]
{
\centering
\includegraphics[width=0.9\textwidth]{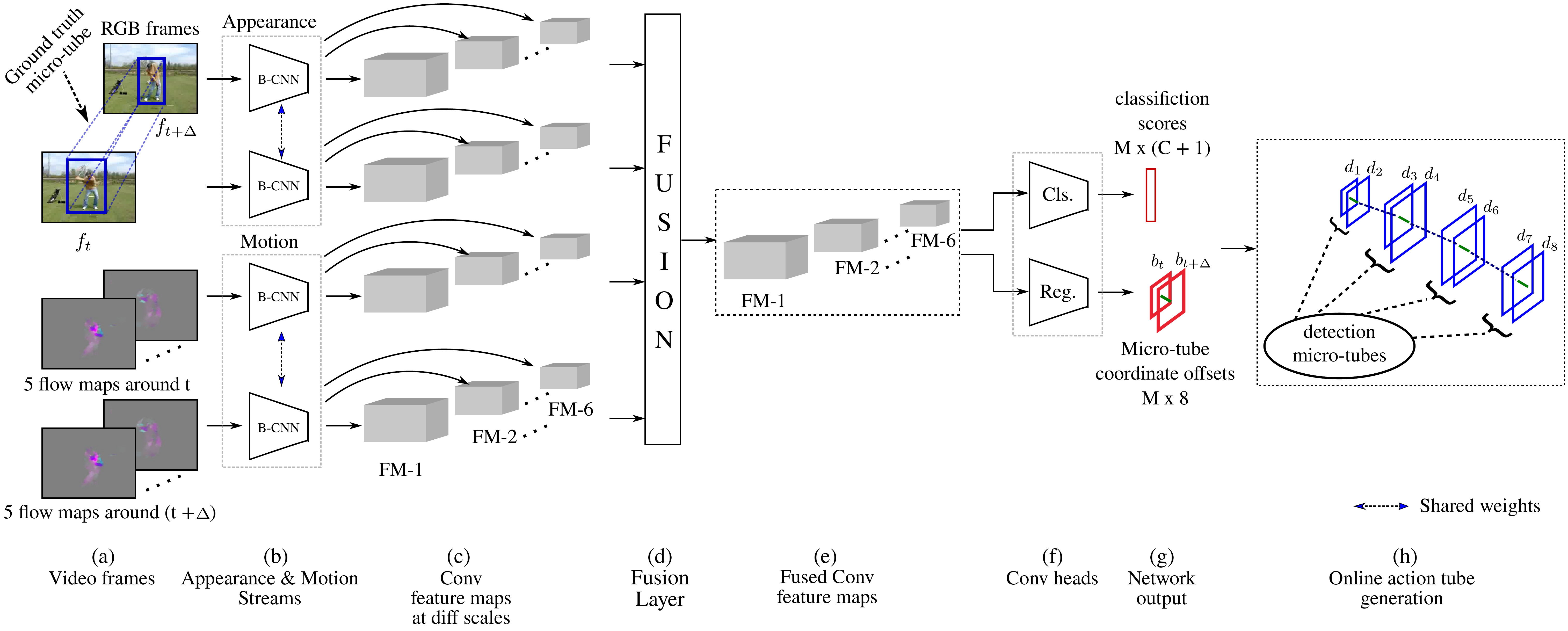}
\caption
{
Overview of the proposed online action detection framework. 
}
\label{fig:overview}
}
\end{figure*} 

\subsection{Rationale}

In this work, we aim to address all such limitations in the existing action detectors.
We propose an action detection framework 
which follows a \emph{video-based} feature representation strategy (in place of a classical {frame-based} one) aimed at encoding spatiotemporal patterns of human actions in videos. 
For this, we build upon and adapt the recent Action Micro Tube (AMTnet)~\cite{saha2017amtnet} deep network architecture. 

The reason for selecting AMTnet over \cite{kalogeiton2017action,hou2017tube} is two-fold.
(a) Firstly, AMTnet works on pairs of video frames for action representation as opposed to \cite{kalogeiton2017action,hou2017tube} which require $k$ consecutive video frames ($k=6$ for \cite{kalogeiton2017action}, $k=8$ for \cite{hou2017tube}). This, as we argued, is computationally expensive and not suitable for real-time applications. 
(b) Secondly, unlike~\cite{kalogeiton2017action,hou2017tube}, AMTnet can work with pairs of frames $(f_t, f_{t+\Delta})$ which are not required to be consecutive (i.e., $\Delta$ does not have to be equal to 1), allowing the system to deal with sparse training annotation ($\Delta$ is large or arbitrary, as in DALY~\cite{daly2016weinzaepfel}).

In this paper, we extend the AMTnet architecture by adding a motion stream, while allowing at the same time:
(1) the end-to-end training of both streams (appearance and motion), and
(2) train-time convolutional feature fusion of RGB and optical flow information.
We show that the latter improves video-mAP over the widely used test-time late fusion
\cite{Georgia-2015a,Weinzaepfel-2015,peng2016eccv,Saha2016,kalogeiton2017action,singh2017online}.
Finally (3), we propose an online algorithm which incrementally builds action tubes by linking micro-tubes in time,
and is shown to outperform existing offline tube generation approaches \cite{Georgia-2015a,Saha2016,peng2016eccv,Weinzaepfel-2015}.
\\
As we show in the following, the resulting action detection framework behaves gracefully when trained on sparse annotations (e.g. $\Delta=\{ 8, 16\}$), achieving similar video-mAPs as models trained on more dense ones (e.g. $\Delta=\{2, 4\}$).

\subsection{Overview of the approach} 

While the proposed online action detection framework (Fig.~\ref{fig:overview}) adapts some of the components of~\cite{saha2017amtnet},
it introduces several major changes to the AMTnet architecture, namely: 
(1) the presence of a motion stream alongside the appearance one (within an end-to-end trainable setup), 
(2) a train time fusion scheme, and
(3) a network architecture geared towards a faster detection speed (in order to meet the online real-time requirements of real world scenarios). 
For tasks (1) and (2), we add a parallel optical flow-based CNN similar to the original RGB CNN, and introduce a new fusion layer designed to merge RGB and optical flow convolutional features (computed at different spatial scales 1,...,6) at training time.
We compute optical flow fields relating successive video frames using~\cite{brox2004high}. 
As for (3), we replace AMTnet's original Faster R-CNN-based~\cite{ren2015faster} network architecture with a comparatively faster SSD~\cite{liu2016ssd} network design.
The SSD architecture minimises computational cost by eliminating the need for a separate region proposal network \cite{ren2015faster}.

The input to the network is a pair of successive (but not necessarily consecutive) RGB video frames $(f_{t}, f_{t+\Delta})$ and the corresponding stacked optical flow maps (Fig.~\ref{fig:overview} \textbf{(a)}). 
These RGB and flow frames are propagated through their respective appearance and motion streams \textbf{(b)}. 
The latter output convolutional feature maps at $6$ different spatial scales \textbf{(c)} which are passed through to a fusion layer \textbf{(d)} where they are merged.
The resulting fused conv feature maps \textbf{(e)} are then passed as inputs to a classification (cls) layer and a regression (reg) layer \textbf{(f)}. 
The cls layer outputs $M \times(C + 1)$ softmax scores for $M$ micro-tubes and $C$ action classes, whereas the reg layer outputs $M \times 8$ bounding box coordinate offsets corresponding to $M$ micro-tubes \textbf{(g)}. 
At test time, the outputs of the cls and reg layers are passed to our online action tube generation algorithm (\textbf{(h)} \S~\ref{sec:methodology:subsec:tube-gen}) which incrementally builds action tubes by linking the detected micro-tubes in time. 

\begin{figure}[!h]
{
\centering
\includegraphics[width=0.4\textwidth]{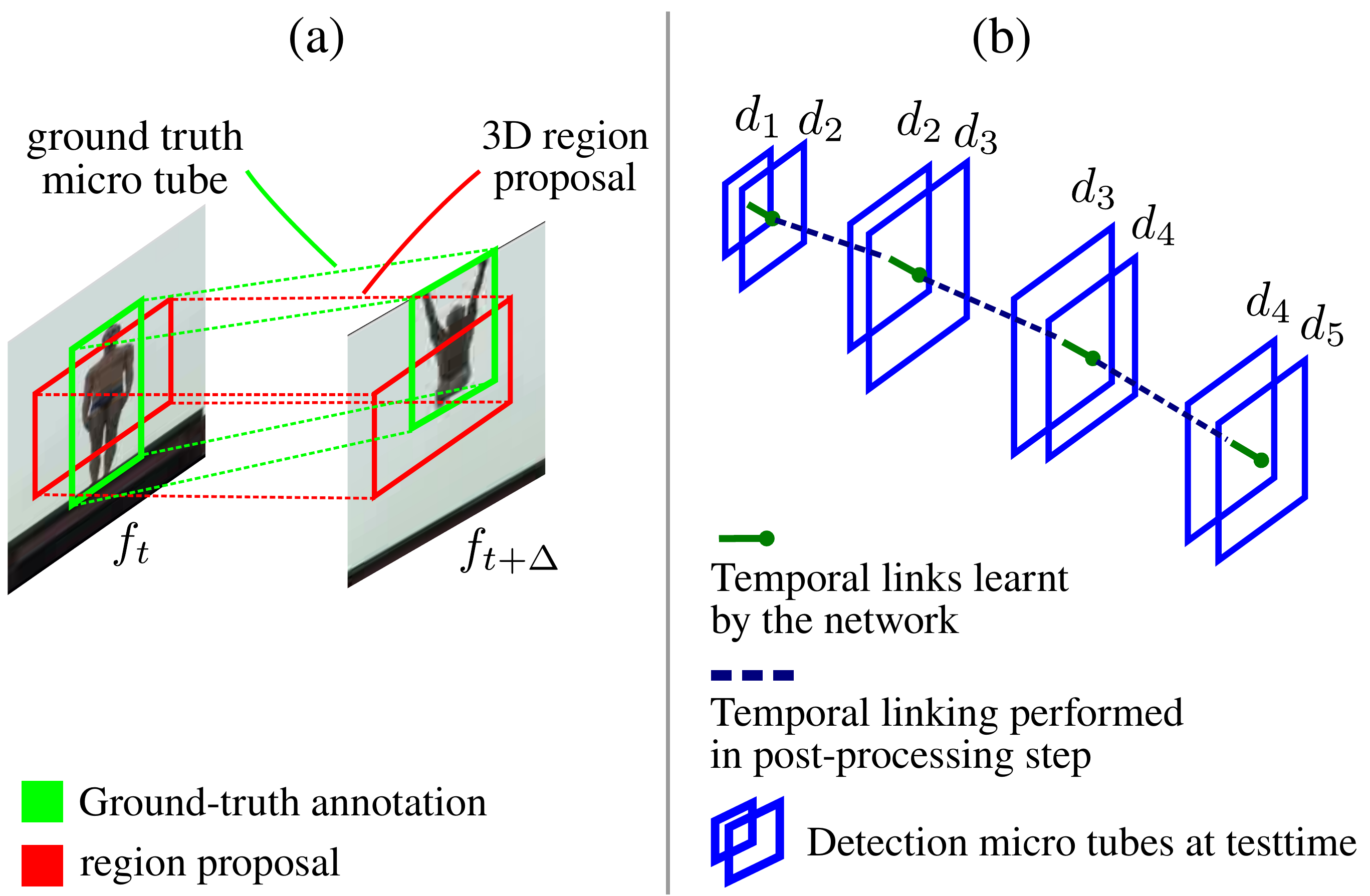}
\caption
{
\textbf{(a)} In AMTnet's video-based action representation a regression layer learns a mapping from a 3D region proposal (in red) to a ground truth micro-tube (in green).
\textbf{(b)} The detected micro-tubes are temporally linked at test time.
}
\label{sec:intro:fig:3dp_mt}
}
\end{figure}

We think useful to elaborate on the concept of action micro-tube (and the related notion of \emph{3D region proposal}) via a simple example, to help the reader understand the action detection framework proposed in this paper.
\\
Consider Figure~\ref{sec:intro:fig:3dp_mt}\textbf{(a)}, where a {``diving''} action spans two successive (but not necessarily consecutive) video frames $f_{t}$ and $f_{t+\Delta}$.
A ground truth action \emph{micro-tube} (i.e., a pair of ground truth bounding boxes belonging to frames $f_{t}$ and $f_{t+\Delta}$, respectively) is shown in green.
One of the best matched \emph{3D region proposals} (i.e., a pair of anchor/default boxes possessing high mean overlap with the corresponding ground truth boxes) is shown in red.
At the time of training, AMTnet's regression layer learns a transformation mapping the 3D region proposal to the ground truth micro-tube.   
\\
Unlike frame-based action detection methods~\cite{Georgia-2015a,Weinzaepfel-2015,Saha2016,peng2016eccv} which output detection bounding-boxes during testing, an action detection model trained on video-based features outputs detection micro-tubes.
Temporally linking detection micro-tubes regressed at test time is faster than linking frame level detections (for more details, please refer to Section~\ref{sec:methodology:subsec:tube-gen}) --   
Figure~\ref{sec:intro:fig:3dp_mt}\textbf{(b)} illustrates the notion.
\\
\SAHA{Note that, unlike AMTnet~\cite{saha2017amtnet} which aims at exploiting only RGB data (without optical flow), the proposed method mainly focuses on learning a spatiotemporal representation (at training time) by leveraging both RGB and optical flow signals. To this end, we propose a better way of fusing the RGB and flow cues by combining the convolutional features during training. In contrast, the exiting action detection methods combine RGB and flow cues at inference time which is indeed suboptimal.
Although, training time fusion has already been investigated for action recognition~\cite{Feichtenhofer_2016_CVPR}, 
we haven't noticed much work exploring fusion at training in the action detection community.
We believe, the proposed new deep architecture with a training time fusion layer would be beneficial for the research community.
}

\subsection{Contributions}
\SAHA{Summarizing, this paper introduces a deep neural network architecture, which we call \emph{``Two-Stream AMTnet''}. 
The key contributions of this work are:
(1) combining AMTnet's 3D proposal architecture with a video-level (as opposed to frame-level~\cite{singh2017online}) online action tube generation algorithm which allows the model to learn stronger temporal features needed for accurate action detection and facilitates running inference online (i.e. incrementally build action tubes in time);
(2) an efficient fusion technique allowing the deep network to merge RGB and optical flow features at training time rather than at test time, thus allowing the network to learn strong sptiotemporal action representations.
In addition, our online tube generation algorithm works in real-time when exploiting only appearance features.}

As a result, we report superior action detection performance over state-of-the-art methods on the standard action detection benchmarks UCF-101-24 and J-HMDB-21.
The source code has been made publicly available at \href{https://github.com/gurkirt/AMTNet}{https://github.com/gurkirt/AMTNet}.
 
\section{Related Work}
\label{sec:related_work} 
Initial attempts to perform space-time action detection were based on either action cuboid hypotheses and sliding-window based approaches~\cite{laptev2007retrieving,cao2010cross,tian2013spatiotemporal}, or
human location-centric based methods~\cite{prest2013explicit,lan2011discriminative,tian2013spatiotemporal,klaser2010human,wang2014video}.
These approaches were limited by the cuboidal shape of their action proposals or their reliance on human location information, respectively.
In opposition, unsupervised 3D action proposals-based methods~\cite{jain2014action,oneata2014spatio,Soomro2015,marian2015unsupervised} 
use a greedy agglomerative clustering of supervoxels for video segmentation, whose major drawback is that they are very expensive and impractical for long duration videos.

\textbf{Frame-based action detection approaches}~\cite{Georgia-2015a,Weinzaepfel-2015,Saha2016,peng2016eccv,singh2017online} demonstrated promising results.
These frameworks mainly leverage object detection algorithms~\cite{girshick-2014,ren2015faster} and the two-stream CNN framework~\cite{simonyan2014two} for action detection.
A frame-based feature representation, however, cannot encode the inter-frame temporal relationships between action regions spanning successive video frames. 
Thus, for linking the frame-level detections in time Viterbi (dynamic programming)~\cite{Georgia-2015a,Saha2016,peng2016eccv} or tracking algorithms~\cite{Weinzaepfel-2015} need to be used.
To put it differently, in frame-based methods the deep network is not in a condition to learn the temporal features essential for accurate action detection, and temporal reasoning is performed by some tube construction stage, in a suboptimal post-processing step. 

\textbf{Video-based action representation approaches}~\cite{kalogeiton2017action,hou2017tube,hou2017end,saha2017amtnet} 
address this issue by learning complex non-linear functions (in the form of CNNs) which map video data (instead of image data) to a high dimensional latent feature space.
Kalogeiton~\etal~\cite{kalogeiton2017action} and Hou~\etal~\cite{hou2017tube,hou2017end}'s models map $K$ video frames to such a latent space, 
whereas Saha~\etal~\cite{saha2017amtnet} only require $2$ successive frames to learn spatio-temporal feature embeddings.
The main advantage of~\cite{saha2017amtnet} over~\cite{kalogeiton2017action,hou2017tube} is that its framework is not constrained to process a fixed set of $K$ frames, but is
flexible enough to select training frame pairs $(f_{t}, f_{t+\Delta})$ at various intervals by varying the $\Delta$ value according to the requirements of a particular dataset
(i.e. a larger $\Delta$ for longer sequences, or a relatively smaller one for shorter video clips). 
This allows the system to leverage the sparse annotation provided with the latest generation action detection datasets, such as AVA~\cite{ava2017gu} and DALY~\cite{daly2016weinzaepfel}.
Another benefit of~\cite{saha2017amtnet}'s work is that it requires less computing time and resources both at train and test time.
Our action detection framework does indeed adapt the network architecture of~\cite{saha2017amtnet} by advancing significant architectural changes, i.e.,
the use of a single-stage base network over a multi-stage backbone, an added motion stream, and a convolutional feature fusion layer (discussed in detail in the next section).
\\
\SAHA{Li~\etal~\cite{li2018recurrent} recently propose two networks - recurrent tubelet proposal (RTP) and recurrent tubelet recognition (RTR) networks for action detection.
RTP generates proposals in a recurrent manner, i.e. estimates the movements of the proposals in the current frame to produce proposals for the next frame.
RTP generates frame level proposals which are linked in time based on their actionness scores to form tubelets. Subsequently, tubelets are temporally trimmed. 
The temporally trimmed tubelets are finally fed into RTR for recognition.
The downsides of~\cite{li2018recurrent} are: requires multi-stage training and inference of RTP and RTR networks; 
can't be used for online setup as RTR needs the entire tublet for classification.
In contrast, our approach follows a single stage training and inference and constructs action tubelets incrementally in an online fashion.
}

\textbf{Online action detection.} So far we have discussed action detetion methods which mostly follow an offline setup, where it is assumed that the entire video clip is available to the system beforehand.
In most real world scenarios, however, this is not the case -- we thus need action detectors which can incrementally detect action instances from a partially observed video in an online fashion.
Early online action prediction works have focussed on dynamic bags of words \cite{Earlyryoo2011human}, structured SVMs \cite{hoai2014max},
hierarchical representations \cite{lan2014hierarchical}, LSTMs and Fisher vectors \cite{de2016online}.
Unlike our framework, however, \cite{Earlyryoo2011human,hoai2014max,lan2014hierarchical} do not perform online action localisation. 
Soomro~\etal~\cite{Soomrocvpr2016} have recently proposed an online method which can predict an action's label and location by observing a relatively smaller portion of the entire video sequence.
However, \cite{Soomrocvpr2016} only works on temporally trimmed videos and not in real-time, due to expensive segmentation.
Singh~\etal~\cite{singh2017online}, on the other hand, have brought forward an online action detection and early label prediction approach which works on temporally untrimmed videos.
Our online action tube generation algorithm is inspired by~\cite{singh2017online}'s approach. The main difference is that the method proposed here is designed to connect in time video-level action micro-tubes, rather than frame-level detection bounding boxes as in \cite{singh2017online}.

\textbf{Spatiotemporal 3D representation}~\cite{carreira2017quo,tran2018closer,xie2018rethinking,nonlocal2018wang} has recently emerged as a dominant force in action detection~\cite{ava2017gu,duarte2018videocapsulenet,girdhar2018video}, amongst others.
Gu~\etal~\cite{ava2017gu}, for instance, combine an inflated 3D (I3D) network~\cite{carreira2017quo} with 2D proposals to exploit the representational power of I3D.
Duarte~\etal~\cite{duarte2018videocapsulenet}, instead, have proposed an interesting 3D video-capsule network for frame-level action segmentation which, however,
does not output action tubes and uses a different evaluation protocol.
\SAHA{Girdhar~\etal~\cite{girdhar2018video} leverage the contextual information (i.e. people and objects around an actor) for improving action detection performance.
They exploit the I3D model and \emph{Transformer} architecture~\cite{vaswani2017attention} to encode contextual information.}
\\
\SAHA{While our goal is to exploit 3D proposals to generate online action tubes, the above methods are inherently offline because of their use of temporal convolution ~\cite{ava2017gu,duarte2018videocapsulenet,girdhar2018video}}.
Additionally, their contribution is mostly on the representation side.
whereas our main contribution is the notion of flexible 3D proposals. For these reasons, they are not directly comparable to our approach. 
We do provide a comparison with~\cite{li2018recurrent}, which uses 3D proposals generated by recurrent networks in an offline setup.
\SAHA{Our approach outperfroms~\cite{li2018recurrent} (\S~\ref{sec:experiment}) on the temporally untrimmed UCF-101-24 dataset
which shows that our 3D proposals are better at encoding the spatiotemporal features than the recurrent tubelet proposals~\cite{li2018recurrent}.} 



\section{Methodology} \label{sec:methodology}

\subsection{Two stream AMTnet architecture} \label{sec:methodology:subsec:net-arch}

\subsubsection{Base network architecture}

We use two stream VGG-16~\cite{simonyan2014two} (B-CNN) as the base network
to implement our appearance and motion CNNs.
The network design of B-CNN is similar to that of an SSD~\cite{liu2016ssd} object detector.
It comprises: 
(1) VGG-16 conv layers, followed by
(2) two conv layers (Conv6 and Conv7), in replacement of the fully connected layers FC-6 and FC-7 of VGG-16, and
(3) $4$ extra convolutional layers (termed Conv8-2, Conv9-2, Conv10-2 and Conv11-2).
The latter have decreasing output dimensions, i.e., they output feature maps characterised by decreasing spatial dimensions. 
This is designed so that B-CNN may learn representations from feature maps at different scales.
The largest dimension ($38 \times 38$) represents the smallest 2D spatial scale, whereas the smallest ($1 \times 1$) corresponds to the largest scale.

\subsubsection{Appearance and motion streams}
 
The appearance and motion streams are comprised of pairs of two parallel B-CNNs with shared weights (Fig.~\ref{fig:overview} (b)).
The input to the appearance stream is a pair of successive RGB video frames ($f_{t}, f_{t+\Delta}$).
For each RGB frame in a pair, the motion stream takes as input 
a stack of 5 optical flow frames, centred around
$t$ and $(t+\Delta)$. Namely, for frame $f_{t}$ the optical flow frames computed in correspondence of $\{f_{t-2}$, $f_{t-1}$, $f_{t}$, $f_{t+1}$, $f_{t+2}\}$ are stacked. A similar sequence is considered for frame $f_{t+\Delta}$.
\\
As a result, the input dimension for these $4$ B-CNNs is $[BS \times D \times H \times W]$, where $BS$ is the training mini-batch size and $[W \times H]$ represent the spatial dimensions of the input video frames ($300 \times 300$ in our case). The depth $D$ is set to $3$ for the B-CNNs working on RGB inputs, to $[3 \times 5]$ for the flow ones.
All four B-CNNs process their respective input frames (either RGB or flow ones) through several convolutional layers (interspersed with ReLu and max-pooling layers),
and output $4$ sets of conv feature maps ($6$ in each set) at different spatial scales (Fig.~\ref{fig:overview} \textbf{(c)}).

\subsubsection{Train time fusion}
 
A strong feature of Two-Stream AMTnet is its train time fusion approach, which makes end-to-end training possible.
Unlike previous action detection works~\cite{Georgia-2015a,Weinzaepfel-2015,peng2016eccv,Saha2016,kalogeiton2017action,singh2017online} which follow a test time late fusion of appearance and motion cues, 
our fusion approach is integrated within the neural network, and thus part of the training optimisation procedure. This allows the network to encode both spatial and temporal features of human actions.
The fusion layer (Fig.~\ref{fig:overview} \textbf{(d)}) accepts as inputs the $4$ sets of conv feature maps (Fig.~\ref{fig:overview} \textbf{(c)}) produced by the appearance and motion streams,
performs the merging of these RGB and flow features and outputs a set of $6$ fused conv feature maps (Fig.~\ref{fig:overview} \textbf{(e)}).

As explained in Section~\ref{sec:exp:subsec:traintime-fusion}, we consider and test two types of feature-level appearance and motion cue fusion techniques -- \emph{element-wise sum} and \emph{feature map concatenation}. 
As the dimension of a concatenated feature map is twice that of each input one, the second strategy requires more parameters for the subsequent classification and regression layers.
In opposition, element-wise sum yields the same dimensionality as in the original feature maps.

\subsubsection{Classification and regression layers}
 
Two-Stream AMTnet's classification and regression layers are implemented using convolutional layers as in~\cite{liu2016ssd}.
The classification ``conv head'' outputs $M \times (C+1)$ class confidence scores for $M$ micro-tubes 
and $C+1$ action classes (one of which is added to represent the `background' class).
The regression conv head outputs $M \times 8$ coordinate offsets 
for $M$ micro-tubes, where $M$ is number of anchor boxes, in our case $8732$.
During testing, we use non-maximum suppression to select the top $k$ detection micro-tubes. 


\begin{figure*}[!th]
  \centering
  \includegraphics[width=0.6\textwidth]{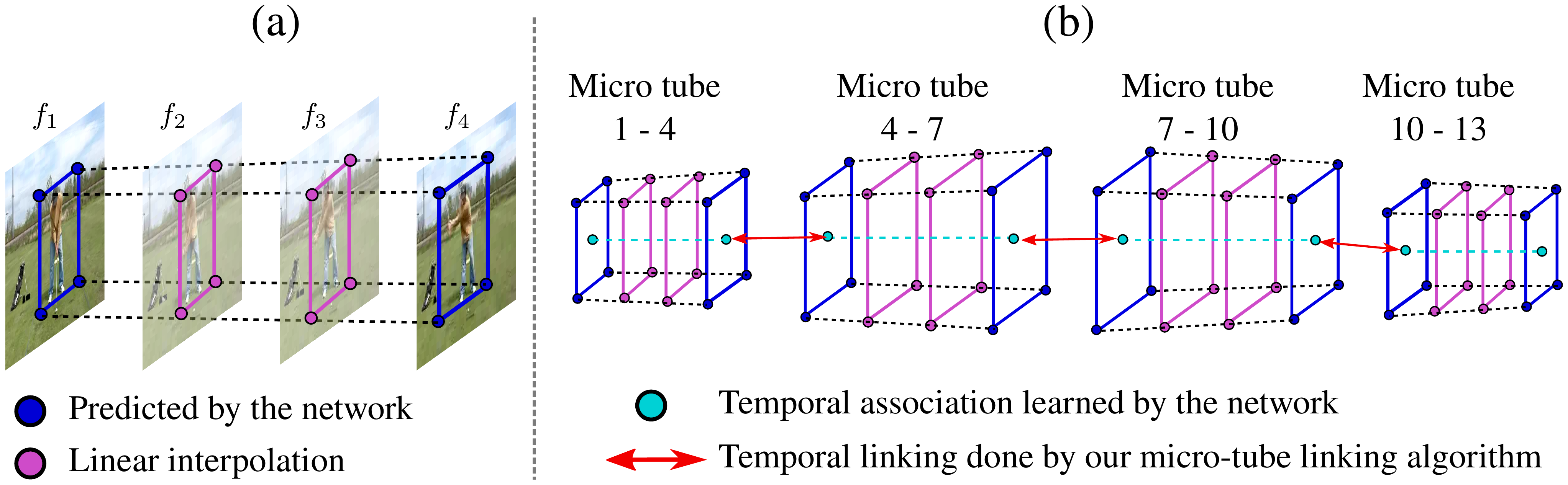}
  \caption
  {
  \textbf{(a)} Action micro-tube generation by linear interpolation of coordinates of predicted bounding boxes. The blue circles denote the $8$ $(x,y)$ coordinate values of the bounding boxes predicted by our network for a pair of successive (but not necessarily consecutive) test video frames.
 The first frame in the pair is indexed by $t=1$, the second by $t=4$.
 We generate the coordinates of the detection bounding boxes (pink circles) for intermediate frames 
 by linear interpolation of the predicted coordinates.
 \textbf{(b)} Action tube generation takes place via the linking of micro-tubes generated as in \textbf{(a)}. Note that, for a video sequence with $T$ frames, our model only needs to perform $T/\Delta$ forward passes -- as a result, the micro-tube linking algorithm (\S~\ref{sec:methodology:subsec:tube-gen}) only needs to connect $T/\Delta -1 $ of these frames. 
  }
  \label{fig:box-intrpl_mt_linking}
\end{figure*}


\subsection{Incremental action tube generation} \label{sec:methodology:subsec:tube-gen}

Another contribution of this work is an online action tube generation algorithm, which processes the micro-tubes predicted by Two-Stream AMTnet (\S~\ref{sec:methodology:subsec:net-arch}).
Indeed, the AMTnet-based tube detection framework proposed in \cite{saha2017amtnet} 
is offline, as micro-tube linking is done recursively in an offline fashion \cite{Georgia-2015a}.
Similarly to Kalogeiton \etal~\cite{kalogeiton2017action}, 
we adapt the online linking method of \cite{singh2017online} 
to link a collection of micro-tubes into a fully-formed action tube. 

Let $B_t$ be the set of detection bounding boxes from a
frame $f_t$. 
Since their method is based on a frame-level detector, Singh \etal~\cite{singh2017online} associate boxes in $B_t$ to boxes in $B_{t+1}$. 
In our case, as Two-Stream AMTnet generates 
micro-tubes $m_t \in M_t \doteq B_t^{1} \times B_{t+\Delta}^{2}$ each spanning
a pair of frames $\{f_t,f_{t+\Delta}\}$ separated by an interval of $\Delta$,
the algorithm needs to link the microtubes associated with the pair $t,t+\Delta$ with those
$m_{t+\Delta} \in M_{t+\Delta} \doteq B_{t+\Delta}^1 \times B_{t+2\Delta}^2$ associated with the following pair of frames 
$\{f_{t+\Delta},f_{t+2\Delta}\}$.
\\
Note that the set of detections $B_{t+\Delta}^{2}$ generated for time $t+\Delta$ by the network when processing the first pair of frames, will in general differ from the set of detections $B_{t+\Delta}^{1}$ for the \emph{same} frame $t+\Delta$ generated by the network when processing the pair $\{f_{t+\Delta},f_{t+2\Delta}\}$.
\\
Micro-tube linking happens by associating elements of $B_{t+\Delta}^2$, coming
from $M_t$, with elements of $B_{t+\Delta}^1$, coming from $M_{t+\Delta}$.
Interestingly, the latter is a relatively easier sub-problem, 
as all such detections are generated based on the same frame, 
unlike the across-frame association problem considered in~\cite{singh2017online}. 
Association is achieved based on both Intersection over Union (IoU) and class score, as 
the tubes are built separately for each class in a multi-label scenario.
For more details, we refer the reader to~\cite{singh2017online}.

\textbf{Bounding box interpolation.} 
At test time, for a input pair of successive video frames ($f_{t}$ and $f_{t+\Delta}$), the final outputs of our Two-Stream AMTnet (\S~\ref{sec:methodology:subsec:net-arch}) are the top $k$ detection micro-tubes and their class confidence scores.
For example, in Figure~\ref{fig:box-intrpl_mt_linking}~\textbf{(a)}, an action micro-tube predicted by the network for an input pair $(f_{1},f_{4})$ ($\Delta = 3$), composed by the two blue bounding boxes highlighted and uniquely determined by 8 coordinate values (blue circles)
is depicted. In practice, the network predicts $M$ such micro-tubes for each input pair, where $M$ is a parameter which depends on the number of anchor locations. 
The predicted micro-tube, however, does not provide bounding boxes for the intermediate frames (in the example $f_{2}$ and $f_{3}$). 
We thus generate detection boxes for all intermediate frames using a simple coordinate wise linear interpolation. 

\section{Datasets and evaluation metrics} 

To evaluate our proposed action detection approach we adopt the following two standard benchmarks.

\textbf{J-HMDB-21}~\cite{J-HMDB-Jhuang-2013} is a spatial action detection dataset (i.e. videos are temporally trimmed as per action duration) which comprises $928$ video clips and $21$ action categories with $3$ train and test splits.
\\
We report evaluation results averaged over the $3$ splits.

\textbf{UCF-101-24}~\cite{soomro-2012} is a spatiotemporal action detection dataset which is a subset of the UCF-101 action recognition dataset.
It consists of $3207$ videos and $24$ action classes.
As standard practice, we report evaluation results on the $1^{st}$ split of UCF-101-24.
Ground-truth bounding boxes for human silhouettes are provided for both the above datasets.

\textbf{Evaluation metrics.} We use the standard \emph{frame-mAP}~\cite{Georgia-2015a,peng2016eccv} and \emph{video-mAP}~\cite{Georgia-2015a,Weinzaepfel-2015,Saha2016,peng2016eccv} (mean average precision) performance measures.
For video-mAP evaluation, we consider the IoU thresholds $\delta = {0.2, 0.5, 0.75, [0.5:0.95]}$.
For the latter interval ($[0.5:0.95]$) we compute the video-mAP at each IoU threshold in the range $[0.5:0.05:0.95]$ with step change $0.05$, to then calculate the average (mean video-mAP).
For frame-mAP computation we set an IoU threshold of $\delta=0.5$ for both J-HMDB-21 and UCF-101-24. 

\section{Experiments} \label{sec:experiment}

In this section, we report different experimental results to support our contributions.
An ablation study of the proposed Two-Stream AMTnet network is presented in Section~\ref{sec:exp:subsec:ablation-study}.
In Section~\ref{sec:exp:subsec:traintime-fusion}, we report the performance gains associated with using our train time fusion strategy as opposed to the widely used test time late fusion.
\\
We then show that Two-Stream AMTnet is robust to sparse training annotation, as action detection performance does not significantly degrade when the network is trained on frame pairs separated by larger $\Delta$ values (\S~\ref{sec:exp:subsec:impact-of-delta}).
Larger $\Delta$s are characteristic of datasets contemplating longer duration videos, and forcibly sparser frame-level bounding-box annotations, such as DALY~\cite{daly2016weinzaepfel} and AVA~\cite{ava2017gu}.
\\
In Section \ref{sec:exp:subsec:impact-of-online-tube},
to attest the effectiveness of our online action tube generation approach, we compare its detection performance to that of state-of-the-art (SOA) approaches. Although our method builds action tubes incrementally (i.e., in an online fashion), it still outperforms offline methods.
In Section~\ref{sec:exp:subsec:train-test-time-comp} we discuss the computing time requirements of the proposed network architecture.
\\

\subsection{Ablation study} \label{sec:exp:subsec:ablation-study}
\begin{table}[t]
\centering
\setlength{\tabcolsep}{4pt}
\caption{
An ablation study of the proposed deep network is presented under different setups. 
The action detection performance (video-mAP and classification accuracy, in \%) of the proposed Two-Stream AMTnet model at different spatiotemporal IoU thresholds ($\delta$) is reported when
training the network on RGB frames only, only optical flow frames only, and when RGB and flow cues are fused using different train and test time fusion techniques.
}
{\footnotesize
\scalebox{1}{
\begin{tabular}{lccccc}
\toprule
IoU threshold $\delta$                  & 0.2   & 0.5   & 0.75  & 0.5:0.95  & Acc \\
\midrule
J-HMDB-21 RGB 				& 59.6	& 58.8 & 44.6	& 37.2     & 52.5 \\
J-HMDB-21 Flow				& 68.0	& 66.1	& 41.3	& 37.5	    & 62.3 \\

J-HMDB-21 late fusion ${}^*$		& 71.7	& 71.2	& 49.7	& 42.5	    & 65.8 \\
J-HMDB-21 concat fusion ${}^+$         	& 73.10& 72.6	& 59.8	& 48.3	    & 68.4 \\
J-HMDB-21 sum fusion ${}^+$            	& 73.5	& 72.8	& 59.7	& 48.1	    & 69.6 \\ 
\midrule
UCF-101-24 RGB 				& 75.8	& 45.3 & 19.9	& 22.0      & 90.0 \\
UCF-101-24 Flow				& 73.7	& 41.6	& 11.3	& 17.1	     & 87.8 \\

UCF-101-24 late fusion ${}^*$		& 79.7	& 49.1	& 19.8	& 22.9	     & 92.3 \\
UCF-101-24 concat fusion ${}^+$        & 78.9 & 49.7 & 22.1 & 24.1  & 91.9  \\
UCF-101-24 sum fusion ${}^+$           & 78.5 & 49.7 & 22.2 & 24.0  & 91.2 \\
\bottomrule
\multicolumn{6}{l}{ ${}^*$ Fusion done at test time.}\\
\multicolumn{6}{l}{ ${}^+$ Fusion done as a part of network training.}\\
\end{tabular}
}
}
\label{table:ablation-study} 
\end{table}
We first present an ablation study to show how each individual network stream (appearance or motion) behaves.
Table~\ref{table:ablation-study} reports the action detection performance of the RGB-only and optical flow-only AMTnet 
on the J-HMDB-21 and UCF-101-24 datasets,
versus that of the overall network, assessed under different fusion strategies.
Both the video-mAP at different spatiotemporal IoU thresholds ($\delta$) and the action classification accuracy are reported for each model.
\\
On J-HMDB-21, the optical flow-based stream displays better video-mAP and classification accuracy, whereas on UCF-101-24 the RGB stream outperforms the motion one.
This indicates that J-HMDB-21 videos are better discriminated by motion cues than by appearance ones. 

\subsection{Impact of train time fusion} \label{sec:exp:subsec:traintime-fusion}

Table~\ref{table:ablation-study} also shows the
effectiveness of training time fusion over the widely used test time late fusion~\cite{Georgia-2015a,Weinzaepfel-2015,peng2016eccv,Saha2016,kalogeiton2017action,singh2017online}.
On both the J-HMDB-21 and UCF-101-24 datasets, the overall action detection performance of train time fusion is superior to than that of test time fusion. Furthermore, element-wise sum fusion exhibits better performance than fusion by feature map concatenation,
especially on UCF-101.

\begin{figure}[!h]
{
\centering
\includegraphics[width=0.4\textwidth]{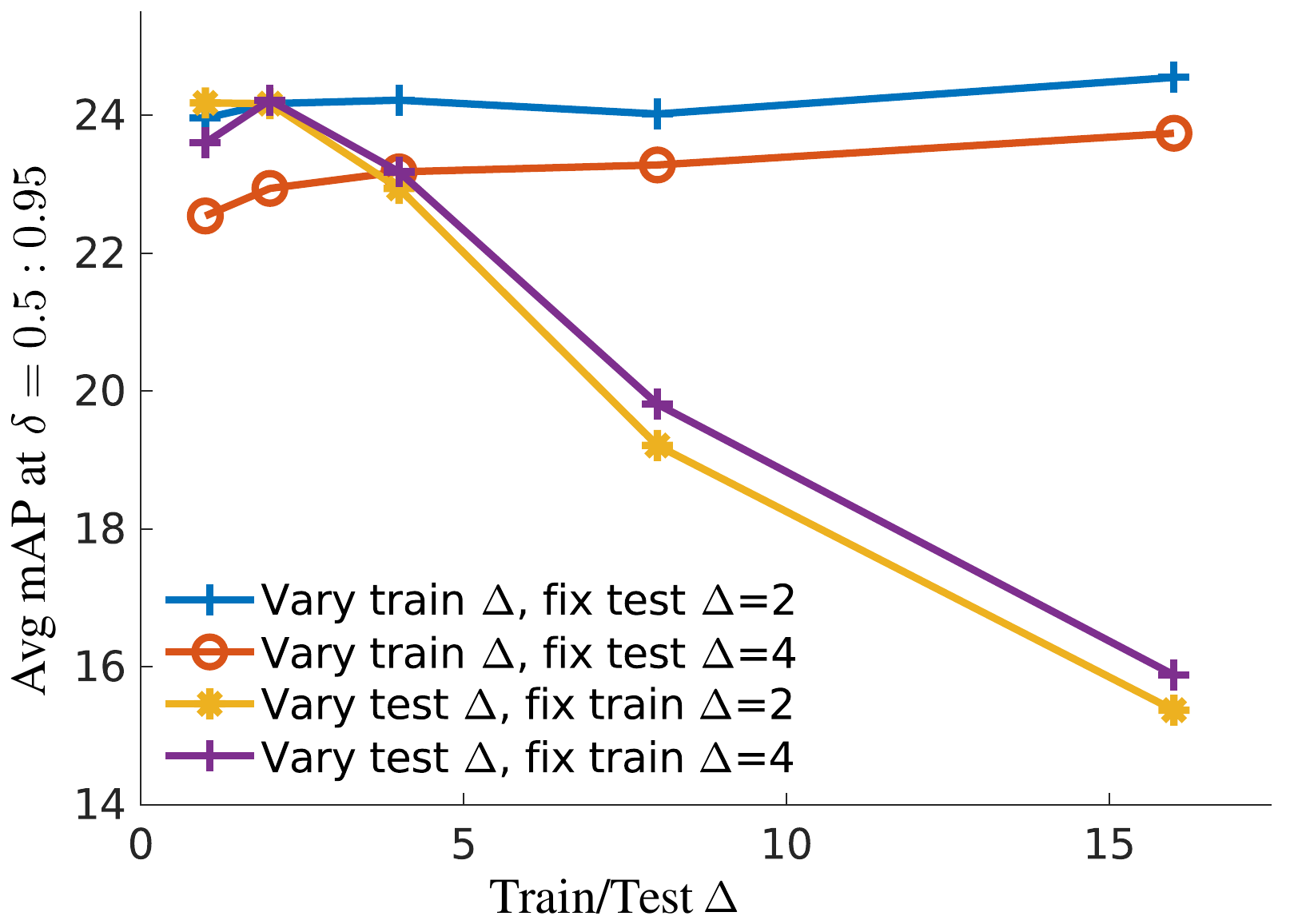}
\caption
{
Video-mAP of the proposed action detection model at different training $\Delta$ values while keeping the test $\Delta$ constant, and viceversa. 
Note that video-mAP does not decrease and actually slightly increases when our model if trained on larger $\Delta$s (e.g., $\Delta=8$ or $16$), 
as opposed to models trained on smaller $\Delta$s (e.g., when $\Delta=1, 2$, and $4$). Varying the test $\Delta$ while keeping train delta constant does instead affect performance, i.e., 
a trade-off exists between detection accuracy and speed.
}
\label{plot:sec:exp:train_delta}
}
\end{figure} 

\subsection{Impact of larger delta values} \label{sec:exp:subsec:impact-of-delta}

As the input to our model is a pair of successive video frames ($f_t, f_{t+\Delta}$) at an inter-frame distance $\Delta$,
it is important to assess how our model reacts to different $\Delta$ values.
In the original AMTnet work, Saha~\etal~\cite{saha2017amtnet} report action detection results using relatively small values of $\Delta$, between $1$ and $3$.
They show that even for consecutive frame pairs (i.e., $\Delta = 1$), AMTnet learns different 
action representations. They do not, however, assess performance at larger $\Delta$~\cite{saha2017amtnet}.

Table~\ref{table:diff-delta-values} and Fig.~\ref{plot:sec:exp:train_delta}, instead, report action detection results for both smaller and larger $\Delta$ values.
\\
For this experiment we selected UCF-101-24, and
trained and tested our detection network using different $\Delta = \{ 1,2,4,8,16\}$ values.
Figure~\ref{plot:sec:exp:train_delta} shows the resulting action detection performance of our framework in terms of video-mAP. 
We first trained distinct Two-Stream AMTnets for the different $\Delta$ values considered.
Each resulting model was then evaluated against different $\Delta$s.
\\
One can note that letting $\Delta$ vary at train time does not affect performance (top two curves).
In fact, Figure~\ref{plot:sec:exp:train_delta} suggests that the network's performance \emph{improves} at wider intervals between annotated training frames. 
These results show the robustness of Two-Stream AMTnet to variations in $\Delta$. 
This is highly beneficial on latest generation action detection datasets (e.g. DALY~\cite{daly2016weinzaepfel} and AVA~\cite{ava2017gu}), which comprise relatively longer video sequences having sparse frame-level bounding box annotations.
At test time, a trade-off exists between performance and speed which can be 
appreciated in the plots (bottom two) where $\Delta$ is varied at test time. 

\begin{table}[t]
\centering
\setlength{\tabcolsep}{4pt}
\caption{
Impact of large $\Delta$ values on action detection performance (video-mAP and classification accuracy, in \%) on the UCF-101-24 dataset.
The video-mAP (at different IoU thresholds) and the action classification accuracy of $5$ different versions of Two-Stream AMTnet, trained on frame pairs separated by different $\Delta$ intervals, are reported.
Very clearly, larger $\Delta$ values such as $8$ and $16$ do not harm action detection performance.
}
{\footnotesize
\scalebox{1}{
\begin{tabular}{lccccc}
\toprule
IoU threshold $\delta$                  & 0.2   & 0.5   & 0.75  & 0.5:0.95  & Acc \\
\midrule
Model $\Delta=1$			& 78.49	& 49.73 & 22.20	& 23.98     & 91.21 \\
Model $\Delta=2$			& 79.73	& 50.47 & 21.66	& 24.18     & 91.32 \\
Model $\Delta=4$			& 79.45	& 48.59 & 21.89	& 23.60     & 91.87 \\
Model $\Delta=8$			& 79.05	& 50.45 & 21.15	& 24.16     & 91.65 \\
Model $\Delta=16$			& 78.77	& 51.17 & 21.89	& 24.46     & 91.65 \\

\bottomrule
\end{tabular}
}
}
\label{table:diff-delta-values} 
\end{table}

\begin{table*}[!t]
    \centering
    \setlength{\tabcolsep}{4pt}
    \caption{Comparison of action detection performance (video-mAP and classification accuracy, in \%) of the proposed Two-Stream AMTnet model with the state-of-the-art on the J-HMDB-21 and UCF-101-24 datasets.}
    {\footnotesize
    \scalebox{1}{
    \begin{tabular}{lccccc|ccccc}
    \toprule
    & \multicolumn{5}{c}{JHMDB-21} & \multicolumn{5}{c}{UCF-101-24} \\ \midrule
    Methods \// $\delta$ =                                   & 0.2           & 0.5               & 0.75              & 0.5:0.95              & Acc \%
                                                             & 0.2           & 0.5               & 0.75              & 0.5:0.95              & Acc \%\\\midrule
    MR-TS Peng~\etal~\cite{peng2016eccv}                     & 74.1          & 73.1              & --                & --                    & --
                                                             & 73.7          & 32.1              & 00.9              & 07.3                  & --\\
    FasterRCNN Saha~\etal \cite{Saha2016}                    & 72.2          & 71.5              & 43.5              & 40.0                  & --
                                                             & 66.6          & 36.4              & 07.9              & 14.4                  & --\\
    SSD + OJLA Behl~\etal~\cite{behl2017incremental}${}^*$   & --            & 67.3              & --                & 36.1                  & --
                                                             & 68.3          & 40.5                & 14.3                & 18.6                    & --\\
    SSD Singh~\etal~\cite{singh2017online}${}^*$             & 73.8          & 72.0              & 44.5              & 41.6                  & --
                                                             & 76.4          & 45.2              & 14.4              & 20.1                  & 92.2 \\
    \midrule    
    AMTnet Saha~\etal~\cite{saha2017amtnet} RGB-only         & 57.7          & 55.3              & --                & --                    & --
                                                             & 63.1          & 33.1              & --                & 10.4                    & --\\
    ACT kalogeiton~\etal~\cite{kalogeiton2017action}${}^*$   & 74.2          & 73.7              & 52.1              & 44.8                  & 61.7 
                                                             & 76.5          & 49.2              & 19.7              & 23.4                  & -- \\
    T-CNN (offline) Hou~\etal~\cite{hou2017tube}             & 78.4          & 76.9              & --                & --                    & 67.2 
                                                             & 47.1          & --                & --                & --                    & -- \\
    RTPR (offline) (VGG) Li~\etal~\cite{li2018recurrent}     & \textbf{82.3} & \textbf{80.5}     & --                                                               & --                    & -- 
                                                             & 76.3                & --                & --                & --                    & -- \\
    \midrule
    Two-Stream AMTnet late fusion${}^*$                      & 71.7          & 71.2              & 49.7              & 42.5                  & 65.8
                                                             & \textbf{79.7}     & 49.1              & 19.8              & 22.9                  & \textbf{92.3} \\
    Two-Stream AMTnet concat fusion${}^*$                    & 73.1          & 72.6              & 59.8              & \textbf{48.3}         & 68.4 
                                                             & 78.9          & \textbf{49.7}     & 22.1              & \textbf{24.1}       & 90.9 \\
    Two-Stream AMTnet sum fusion${}^*$                       & 73.5          & 72.8              & \textbf{59.7}     & 48.1                 & \textbf{69.6} 
                                                             & 78.5          & \textbf{49.7}     & \textbf{22.2}     & 24.0                 & 91.2 \\
    \bottomrule
    \multicolumn{11}{l}{ ${}^*$ online methods}\\
    \end{tabular}
    }
    }
    \label{table:comp-soa} 
\end{table*}

\subsection{Comparison with the state of the art} \label{sec:exp:subsec:impact-of-online-tube}
Importantly, we can 
quantitatively demonstrate that, despite being designed for online detection, Two-Stream AMTnet outperforms most of the offline methods on both J-HMDB-21 and UCF-101-24, as illustrated in 
Table~\ref{table:comp-soa}.
Our \emph{sum fusion} approach exhibits significant improvements on video-mAP at larger IoU thresholds ($\delta = 0.75$ and $0.5$:$0.95$) on both datasets.
On J-HMDB-21, it shows a video-mAP gain of \textbf{7.6\%}, \textbf{3.3\%} at IoU thresholds of $0.75$ and $0.5$:$0.95$, respectively.
On UCF-101-24, it improves video-mAP by \textbf{2.5\%} and \textbf{0.6\%} at IoU thresholds of $0.75$ and $0.5$:$0.95$, respectively.
Only on J-HMDB-21 \cite{hou2017tube} shows better video-mAPs for relatively smaller detection overlap values ($\delta = 0.2, 0.5$). The authors do not report results on higher $\delta$s ($0.75$ and $0.5$:$0.95$), for which our method otherwise achieves superior detection results.

\SAHA{Interestingly, our model outperforms all existing works on the temporally trimmed UCF-101-24 dataset. As UCF-101-24 has untrimmed videos where action instances can present at any spatial or temporal locations, detection becomes more challenging as compared to the J-HMDB-21 in which video clips are temporally trimmed and they have relatively shorter durations.
On J-HMDB-21, our close competitors are~\cite{peng2016eccv,kalogeiton2017action,hou2017tube,li2018recurrent}.
Peng~\etal~\cite{peng2016eccv} follow an expensive multi-stage training process in which the appearance and the motion streams are trained independently. 
In addition, their training and inference is based on a multi-scale setup, i.e., one video frame is re-scaled several times at different resolutions, and then used for training and inference.
It is quite apparent that \cite{peng2016eccv} is way more expensive than our single-stage end-to-end trainable architecture which does not requires multiple rescaled versions of the input frame.
Our approach outperforms Kalogeiton~\etal~\cite{kalogeiton2017action} on the higher IoU thresholds (i.e. $0.75$, $[0.5:0.95]$) for J-HMDB-21 dataset attesting the robustness of our model towards action localization.
Furthermore, \cite{kalogeiton2017action} is limited by the assumption that it needs an input of $K$ frames (where $K=8$) resulting an expensive forward pass during training and inference.
Quite the contrary, our model needs only a pair of frames during both training and testing.
Hou~\etal~\cite{hou2017tube}'s offline approach shows relatively better results than our online method on the temporally trimmed J-HMDB-21 video clips. 
However, their model performs badly on the temporally untrimmed UCF-101-24 videos showing a lack of spatiotemporal localization capability of their model.
Similarly, Li~\etal~\cite{li2018recurrent} reported superior video mAPs on J-HMDB-21, but for UCF-101-24, our model achieved the best performance.
Besides, \cite{li2018recurrent} requires a multi-stage training and it can't be used in an online setup as already discussed in Section~\ref{sec:related_work}.
}

\subsection{Test Time Detection Speed}\label{sec:exp:subsec:train-test-time-comp}
\label{sec:detection-speed}

We conclude by showcasing the online and real-time capabilities of our approach, bulding on our online tube generation algorithm (\S~\ref{sec:methodology:subsec:tube-gen}), derived from that first presented in Singh~\etal ~\cite{singh2017online}.

The crucial point of our modified algorithm is the speed of
AMTnet's forward pass, since the network processes two frames ($\Delta$ apart) simultaneously.
To this extent, we measure the average time taken for 
a forward pass for a batch size of 1, as compared to 8 in~\cite{singh2017online}. 
A single forward pass takes 46.8 milliseconds to process one test example, 
showing that it can be run in almost real-time at 21fps with two streams on a single 1080Ti GPU. A single stream can be processed at 33fps. 
\\
The processing of a single appearance stream is thus completely online and real-time, as is the case in Singh \etal~\cite{singh2017online}. The two-stream pipeline, however, is slightly slower and has a bottleneck in terms of optical flow computation time. In this work, we use dense optical flow estimation~\cite{brox2004high}, which is relatively slow. Just as in~\cite{singh2017online}, however,
one can always switch to real-time optical flow estimation~\cite{kroeger2016fast}, paying only a small drop in performance. 
One can improve speed even further by testing Two-Stream AMTnet with $\Delta$ equal to $2$ or $4$, for instance, achieving a speed improvement of $2\times$ and $4\times$, respectively.


 
\section{Conclusion}

We presented a novel deep learning framework for action detection which leverages recent advancements on video based action representation~\cite{saha2017amtnet} 
and incremental action tube generation~\cite{singh2017online}.
A number of contributions were highlighted.

Firstly, the proposed framework can efficiently learn spatiotemporal action representations from video data, 
thus sidestepping the issues associated with frame-based approaches~\cite{Georgia-2015a,Weinzaepfel-2015,peng2016eccv,Saha2016}.
Secondly, the approach is the first to incorporate a~\emph{train time fusion} strategy, as opposed to the traditional~\emph{test time fusion}~\cite{Georgia-2015a,Weinzaepfel-2015,peng2016eccv,Saha2016}. 
This allows the network to learn the complementary contributions of appearance and motion cues at training time, and makes the end-to-end joint optimisation of appearance and motion streams possible. The reported experimental results attested the efficacy of our new train time fusion approach over the traditional test time fusion scheme.
\\
As a third contribution, our framework can build action tube incrementally in an online fashion at a faster detection speed ($33$/$21$ fps), 
thus paving the way for its application to real-time problems such as self-driving cars and drones, surgical robotics, or human-robot interaction, among others.
\\
Finally, the proposed action detection framework addresses the paucity of frame-level bounding box annotations which characterises the latest generation action detection benchmarks 
(i.e., DALY~\cite{daly2016weinzaepfel} and AVA~\cite{ava2017gu}) by building on an AMTnet-based network architecture, which only requires pairs of frames at training time, as opposed to its 
close competitors~\cite{kalogeiton2017action,hou2017tube}. In other words, our model supports both sparse and dense annotation, whereas~\cite{kalogeiton2017action,hou2017tube} can only exploit dense annotation, which is very expensive to obtain for large-scale video datasets.

Our method achieves superior performances compared to the previous state-of-the-art in online action detection, while outperforming most if not all top offline competitors, in particular at high detection overlap.
Its combination of high accuracy and fast detection speed at test time is promising and suitable for real-time applications.

\ifCLASSOPTIONcompsoc
  \section*{Acknowledgments}
\else
  \section*{Acknowledgment}
\fi

This project has received funding from the European Union's Horizon 2020 research and innovation programme under grant agreement No. 779813 (SARAS)

\ifCLASSOPTIONcaptionsoff
  \newpage
\fi


\bibliographystyle{IEEEtran}
\bibliography{ref}


\begin{IEEEbiography}[{\includegraphics[width=1in,height=1.25in,clip,keepaspectratio]{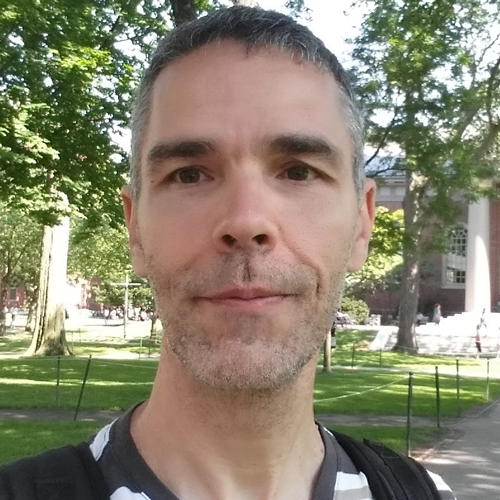}}]{Fabio Cuzzolin}
received in 1997 a laurea  magna  cum  laude in  Computer  Engineering from the University of Padua, Italy. 
He was  awarded  a  Ph.D.  by  the  same  institution in 2001 for the thesis “Visions of a generalized probability theory”. After conducting research at Politecnico di Milano, the Washington University
in St Louis, UCLA and INRIA Rhone-Alpes.
He is currently a Professor and the head of the Visual Artificial Intelligence Laboratory within the School of Engineering, Computing and Mathematics at Oxford Brookes University (UK).
His research interests span artificial intelligence, machine learning and computer vision. He is a world expert in the theory of belief functions, to which he contributed with a general
geometric approach to uncertainty measures to be published soon in a  Springer  monograph.  
He  is  currently  the  author  of  about  80  peer reviewed  publications.  His  work  has  won  a  number  of  awards.
He  is Associate Editor of the IEEE Transactions on Human-Machine Systems and the IEEE Transactions on SMC-C.
\end{IEEEbiography}
\vspace{-5mm}
\begin{IEEEbiography}[{\includegraphics[width=1in,height=1.25in,clip,keepaspectratio]{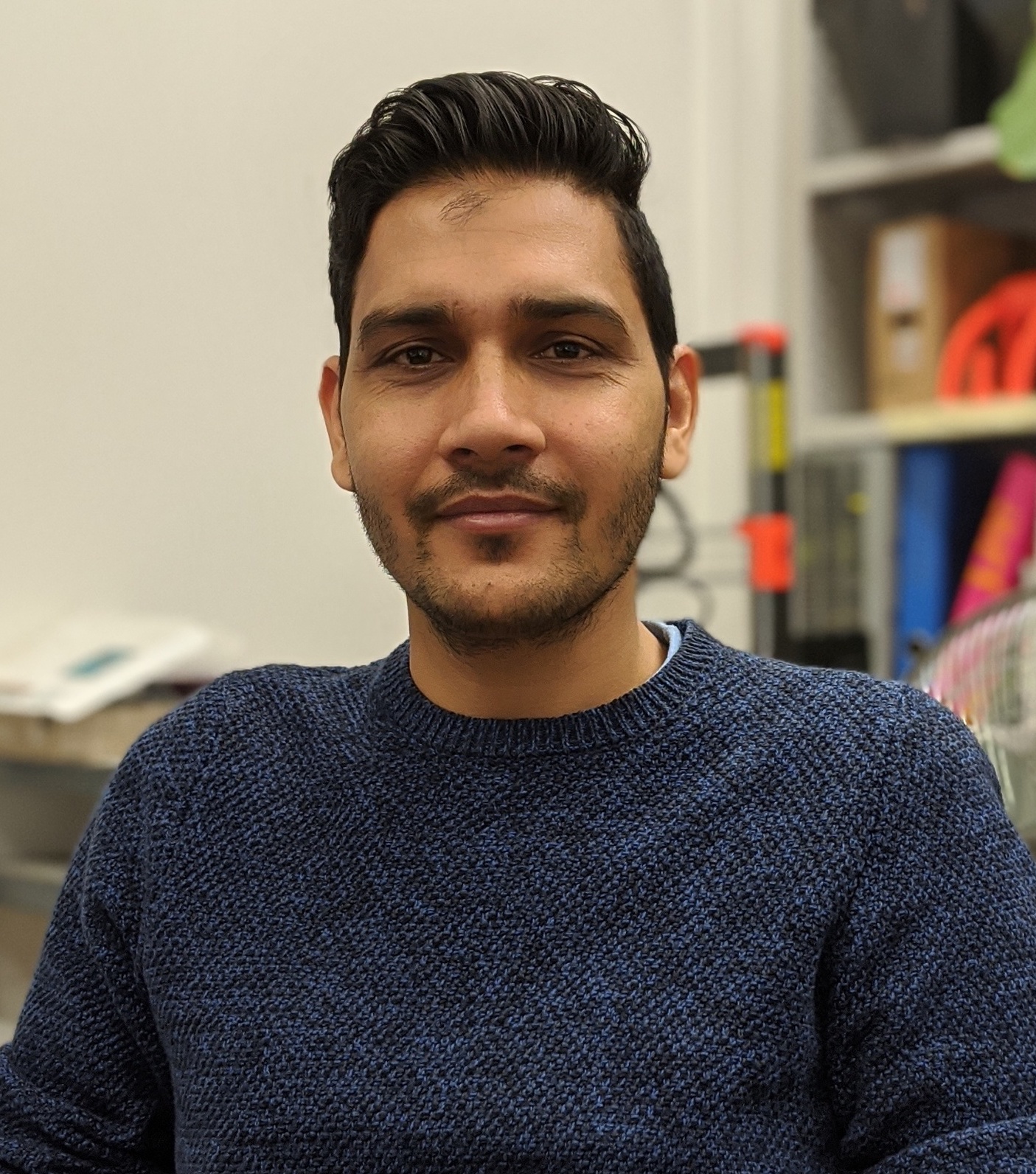}}]{Gurkirt Singh}is a post doctoral research fellow at the Computer Vision Lab (CVL), ETH Zurich, Switzerland.
He works with Professor Luc Van Gool at CVL.
Before joining ETH Zurich, he was a Research Associate (RA) at the AI and CV group, Oxford Brookes University (UK), where he received his PhD degree.
He was advised by Professor Fabio Cuzzolin.
Earlier, he was a research engineer for two years in imaging and computer vision group at Siemens research India.
He graduated from MOSIG master program at Grenoble-INPG (School ENSIMAG) with specialization in Graphics Vision and Robotics.
He completed his master's thesis under the supervision of Dr. Georgios Evangelidis and Dr. Radu HORAUD at INRIA, Grenoble.
He received his Bachelor of Technology degree in Electronics and Instrumentation Engineering from VIT University, Vellore, India.
\end{IEEEbiography}
\vspace{-5mm}
\begin{IEEEbiography}[{\includegraphics[width=1in,height=1.25in,clip,keepaspectratio]{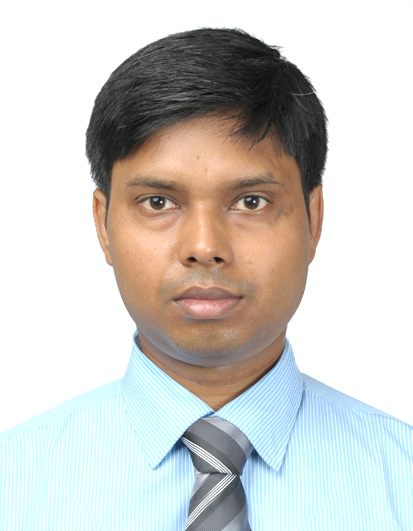}}]{Suman Saha}
is a post doctoral research fellow at the Computer Vision Lab (CVL), ETH Zurich, Switzerland.
He works with Professor Luc Van Gool at CVL.
Before joining ETH Zurich, he was a Research Associate (RA) at the AI and CV group, Oxford Brookes University (UK).
He received his PhD degree in Computer Science and Mathematics under the supervision of Professor Fabio Cuzzolin at Oxford Brookes University.
Suman received his Master's degree (in Computer Science) from University of Bedfordshire (UK).
His Master thesis supervisors were Dr. Ashutosh Natraj and Dr. Sonia Waharte (both were postdoc in the Department of Computer Science, University of Oxford). 
Earlier,  he received his 3 years Polytechnic Diploma Engineering degree in Computer Science from Siddaganga Polytechnic College, India. 
\end{IEEEbiography}







\end{document}


%

\title{Two-Stream AMTnet for Action Detection \\ Supplementary Materials}

\author{Suman~Saha$^{*}$,~Gurkirt~Singh$^{*}$~and~Fabio~Cuzzolin

}

%
%

\markboth{IEEE Transaction on Pattern Analysis and Machine Intelligence,~Vol.~X, No.~Y, 20XX}%
{Shell \MakeLowercase{\textit{et al.}}: Bare Demo of IEEEtran.cls for Computer Society Journals}
%



\IEEEtitleabstractindextext{%

}

\maketitle

\IEEEdisplaynontitleabstractindextext

%
\IEEEpeerreviewmaketitle

\onecolumn
\section{Additional Experimental Results and Source Code Link}
To ablate the proposed Two-Stream AMTnet mode,
we report per class video APs (for 3 diff. fusion methods) in Table \ref{table:class_aps_02} and \ref{table:class_aps_05} 
at spatio-temporal IoU thresholds 0.2 and 0.5 respectively.
The source code has been made publicly available at \href{https://github.com/gurkirt/AMTNet}{https://github.com/gurkirt/AMTNet}.
\begin{table*}[!t]
\centering
\setlength{\tabcolsep}{4pt}
\caption{
  An ablation study of the proposed Two-Stream AMTnet model with three different fusion methods - late, concat and sum fusions.
  Per class video-APs (in \%) for 24 action categories of UCG-101-24 dataset at spatiotemporal IoU threshold 0.2.
  }

{\footnotesize
\scalebox{1}{
\begin{tabular}{lcccccc}
\toprule
Actions             & Basketball            & BasketballDunk        & Biking            & CliffDiving           & CricketBowling            & Diving   \\ \midrule 
Late fusion         & 51.38                 & 83.01                 &  \RD{82.06}       & 83.21                 &  \RD{72.21}               & 100    \\
Concat fusion       & \RD{56.01}            &  \RD{93.09}           & 80.63             &  \RD{85.99}           & 59.09                     & 100    \\
Sum fusion          & 54.19                 & 90.05                 & 79.58             & 83.14                 & 56.5                      & 100     \\  \\ \midrule

Actions             & Fencing               & FloorGymnastics       & GolfSwing         & HorseRiding           & IceDancing                & LongJump   \\ \midrule 
Late fusion         & 88.9                  & 99.2                  &  \RD{77.55}       &  \RD{96.15}           &  \RD{86.09}               &  \RD{85.18}     \\
Concat fusion       &  \RD{93.35}           &  \RD{99.62}           & 68.66             & 96.08                 & 58.11                     & 79.54     \\
Sum fusion          & 93.04                 & 99.25                 & 70.3              & 96.08                 & 57.11                     & 81.62     \\  \\ \midrule

Actions             & PoleVault             & RopeClimbing          & SalsaSpin         & SkateBoarding         & Skiing                    &  Skijet   \\ \midrule 
Late fusion         & 92.54                 & 98.36                 & 45.39             & 86.1                  & 82.09                     &  \RD{89.98}     \\
Concat fusion       &  \RD{95.46}           & 98.62                 & 63.47             & 87.7                  & 85.6                      & 89.13     \\
Sum fusion          & 93.86                 &  \RD{98.82}           &  \RD{64.07}       &  \RD{87.84}           & \RD{85.89}                & 89.58   \\  \\ \midrule

Actions             & SoccerJuggling        & Surfing               & TennisSwing       & TrampolineJumping     & VolleyballSpiking         &  WalkingWithDog   \\ \midrule 
Late fusion         &  \RD{92.29}           &  \RD{72.44}           & 43.2              &  \RD{62.98}           & 50.53                     &  \RD{90.79}  \\
Concat fusion       & 85.29                 & 70.52                 & 47.05             & 59.55                 & 51.58                     & 89.4   \\
Sum fusion          & 87.45                 & 67.89                 &  \RD{47.92}       & 58.26                 &  \RD{52.28}               & 89.07    \\  

\bottomrule
\end{tabular}
}
}
\label{table:class_aps_02}
\end{table*}

\begin{table*}[!t]
\centering
\setlength{\tabcolsep}{4pt}
\caption{
An ablation study of the proposed Two-Stream AMTnet model for three different fusion methods - late, concat and sum fusions.
Per class video-APs (in \%) for 24 action categories of UCG-101-24 dataset at spatio-temporal IoU threshold 0.5.
}
{\footnotesize
\scalebox{1}{
\begin{tabular}{lcccccc}
\toprule
Actions             & Basketball    & BasketballDunk    & Biking        & CliffDiving   & CricketBowling    & Diving   \\ \midrule 
Late fusion         & 0.15          &  \RD{3.32}        & 58.04         & 44.12         & 1.71              & 29.19    \\
Concat fusion       &  \RD{0.78}    & 3.25              &  \RD{63.91}   & 44.1          & 2.13              & 42.28     \\
Sum fusion          & 0.71          & 2.32              & 63.57         &  \RD{44.91}   &  \RD{2.14}        &  \RD{46.24}    \\  \\ \midrule

Actions             & Fencing       & FloorGymnastics    & GolfSwing    & HorseRiding   & IceDancing        & LongJump   \\ \midrule 
Late fusion         & 75.23         & 96.49              &  \RD{48.04}  & 89.46         &  \RD{66.05}       &  \RD{64.1}     \\
Concat fusion       & 81.23         &  \RD{99.62}        & 42.86        & 91.18         & 42.63             & 55.94   \\
Sum fusion          &  \RD{82.2}    & 96.26              & 42.21        &  \RD{94.04}   & 44.72             & 58.76     \\  \\ \midrule

Actions             & PoleVault     & RopeClimbing       & SalsaSpin     & SkateBoarding & Skiing            &  Skijet   \\ \midrule 
Late fusion         & 42.2          & 98.36              & 2.55          & 86.1          & 70.89             & 67.56     \\
Concat fusion       & 52.12         & 98.62              &  \RD{10.27}   & 87.7          &  \RD{82.58}       &  \RD{68.45}   \\
Sum fusion          &  \RD{52.71}   &  \RD{98.82}        & 10.17         &  \RD{87.84}   & 78.46             & 62.74     \\  \\ \midrule

Actions             & SoccerJuggling & Surfing           & TennisSwing      & TrampolineJumping         & VolleyballSpiking         &  WalkingWithDog   \\ \midrule 
Late fusion         &  \RD{87.78}    & 48.97             &  \RD{0.59}       &  \RD{36.96}               &  \RD{0.05}                & 60.28 \\
Concat fusion       & 80.49          &  \RD{51.91}       & 0.29             & 27.77                     & 0.03                      &  \RD{64.56}    \\
Sum fusion          & 83.78          & 46.56             & 0.33             & 29.97                     & 0                         & 64.07  \\  

\bottomrule
\end{tabular}
}
}
\label{table:class_aps_05}
\end{table*}